\pdfoutput=1

\documentclass[11pt]{article}

\usepackage{acl}

\usepackage{times}
\usepackage{latexsym}

\usepackage[T1]{fontenc}

\usepackage[utf8]{inputenc}

\usepackage{microtype}

\usepackage{inconsolata}

\usepackage{algorithm}
\usepackage{algorithmic}
\usepackage{newfloat}
\usepackage{listings}
\usepackage{inconsolata}
\usepackage{enumitem}
\usepackage{bbm}
\usepackage{algorithm}
\usepackage{algorithmic}
\usepackage{multirow}
\usepackage{multicol}
\usepackage{graphicx}
\usepackage{xcolor}
\usepackage{color}
\usepackage{colortbl}
\usepackage{float}
\usepackage{subcaption}
\usepackage{amsmath}
\usepackage{appendix}

\def \st{\;\text{s.t.}\;}

\newcommand{\bh}{\mathbf{h}}

\DeclareMathOperator*{\argmax}{argmax}

\title{Knowledge Graph-Enhanced Large Language Models via Path Selection}

\author{Haochen Liu, Song Wang, Yaochen Zhu, Yushun Dong, Jundong Li\\
  University of Virginia\\  \texttt{\{sat2pv,sw3wv,uqp4qh,yd6eb,jundong\}@virginia.edu}\vspace{0.15in}
  }

\begin{document}
\maketitle
\begin{abstract}

Large Language Models (LLMs) have shown unprecedented performance in various real-world applications. However, they are known to generate factually inaccurate outputs, a.k.a. the hallucination problem. In recent years, incorporating external knowledge extracted from Knowledge Graphs (KGs) has become a promising strategy to improve the factual accuracy of LLM-generated outputs. Nevertheless, most existing explorations rely on LLMs themselves to perform KG knowledge extraction, which is highly inflexible as LLMs can only provide binary judgment on whether a certain knowledge (e.g., a knowledge path in KG) should be used. In addition, LLMs tend to pick only knowledge with direct semantic relationship with the input text, while potentially useful knowledge with indirect semantics can be ignored. In this work, we propose a principled framework KELP with three stages to handle the above problems. Specifically, KELP is able to achieve finer granularity of flexible knowledge extraction by generating scores for knowledge paths with input texts via latent semantic matching. Meanwhile, knowledge paths with indirect semantic relationships with the input text can also be considered via trained encoding between the selected paths in KG and the input text. Experiments on real-world datasets validate the effectiveness of KELP.\footnote{Our code is available at \url{https://github.com/HaochenLiu2000/KELP}.}

\end{abstract}

\section{Introduction}\label{introduction}

Recently, Large Language Models (LLMs) such as ChatGPT~\cite{gpt} and LLaMa~\cite{llama} have shown exceptional performance, such as unprecedented reasoning capabilities~\cite{increasesize}, across various NLP tasks~\cite{QA,translation,zhu2024collaborative,wang2021hierachical}. 
However, in scenarios where certain new knowledge beyond the scope of training corpus is required, current LLMs are usually criticized for generating factually inaccurate outputs~\cite{factualerror1,factualerror2,factualerror3, wang2023noise}. As a consequence, it becomes imperative to develop effective and efficient techniques for incorporating new knowledge into pretrained LLMs.



To facilitate the incorporation of new knowledge in LLMs, extracting external knowledge from Knowledge Graphs (KGs) (known as KG-Enhanced LLMs~\cite{survey1}) has become a promising way to improve the factual accuracy of LLM outputs~\cite{survey2,survey1,wu2024usable}. Here, the structure of KGs plays a critical role, since the relationship between entities can effectively contribute to novel knowledge required by various tasks, such as multi-hop reasoning~\cite{structgpt}. 
In general, there are two mainstreams to achieve KG-Enhanced LLMs. The first mainstream incorporates new knowledge for LLMs during their training phase, which is usually achieved by designing new training objectives or tasks~\cite{ernie,kepler}. Nevertheless, these techniques typically require significant computational resources.
On the contrary, the second mainstream incorporates new knowledge for LLM during the inference phase, where this new knowledge is incorporated by prompt engineering, that is, designing new prompts to include the triplets (head, relation, tail) derived from KGs~\cite{kggpt}. Usually, prompt engineering stands out as the most computationally efficient approach to incorporate new knowledge for LLMs, as new information can be directly introduced together with the text input without an additional training process.




\begin{figure*}[!t]
  \centering 
  \includegraphics[width=01\linewidth]{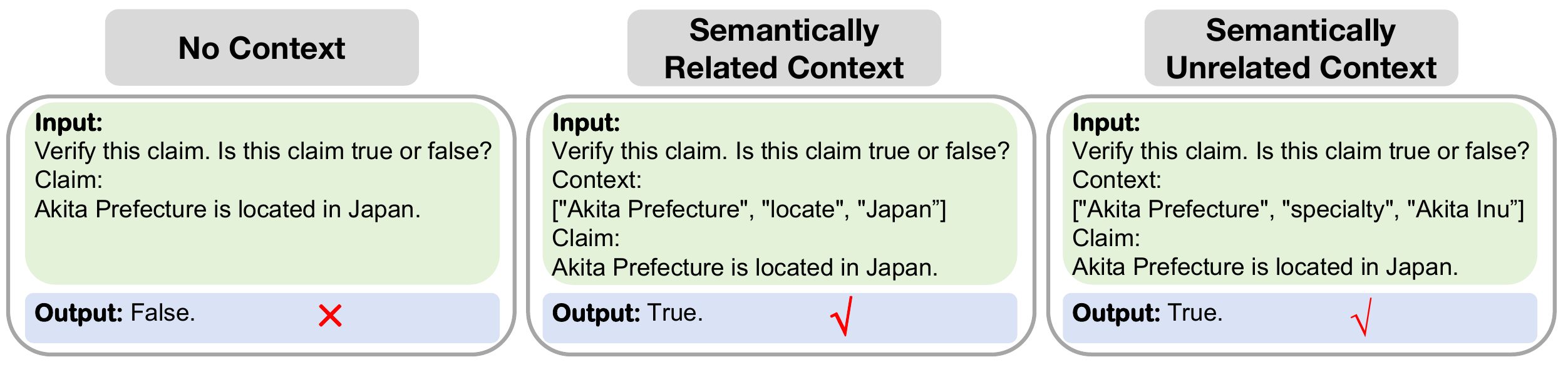}
  \caption{An example of the phenomenon 
 that semantically unrelated contexts in the input prompts can possibly contain important knowledge to correct/improve the generation of large language models. In this example, there exist potential relationships between "Japan" and "Akita Inu" that are challenging to directly identify and capture.}
  \label{example}
\end{figure*}

Nevertheless, integrating new knowledge for LLMs with prompt engineering bears two significant disadvantages despite its effectiveness and efficiency. First, most existing methods solely rely on LLMs to identify relevant triplets~\cite{kggpt}. However, LLMs can only provide binary outputs on whether a certain instance of knowledge (e.g., a path in a KG) should be used or not~\cite{structgpt}. As a consequence, existing approaches bear low flexibility due to their coarse granularity in determining to what extent a certain knowledge is useful.
Second, solely using LLMs to select and incorporate instances of knowledge is usually overwhelmed by the knowledge that has direct semantic relationships with the input text. However, the knowledge with indirect semantic relationships could also help LLMs achieve factually accurate outputs. 
More specifically, instances of knowledge with indirect semantic relationship to the input text can also help LLMs generate factually accurate outputs. 
We refer to these instances as \textit{potentially impactful knowledge}, and an example is presented in Figure~\ref{example}. Such a phenomenon can be attributed to certain potential relationships between entities and relations contained in the training corpus of LLMs, while it can be challenging for both humans and LLMs themselves to perceive directly. Therefore, it is usually difficult for existing approaches to capture this nuanced, potentially impactful knowledge that can effectively improve the LLM outputs solely based on the selection of knowledge instances from LLM themselves.

To properly handle the problems discussed above, we introduce a novel approach for KG-Enhanced LLMs, i.e., KELP (\textbf{K}nowledge Graph-\textbf{E}nhanced Large \textbf{L}anguage Models via \textbf{P}ath Selection), aiming to flexibly capture potentially impactful knowledge as in-context facts to improve the factual accuracy of the LLM outputs given input texts. In particular, KELP consists of three key components: \textit{(i)} Knowledge path extraction, \textit{(ii)} Sample encoding, and \textit{(iii)} Fine-grained path selection. Specifically, we first extract knowledge paths from KG based on the entities identified in the input texts as the candidate knowledge. Subsequently, we train a path-text encoder to encode the \textit{indirect connections} between input texts and the knowledge paths extracted from KG, with similarity defined on the latent semantic space, such that whether an instance of knowledge (represented by a path in the KG) is potentially useful for a certain input text (i.e., is the potentially impactful knowledge) can be quantitatively measured. Based on the latent similarity score, two \emph{coverage rules} are introduced to further refine the selected paths with high flexibility.
Through these meticulously designed steps,
KELP strives to flexibly capture potentially impactful knowledge with fine granularity (based on quantitative scores) to refine LLM outputs. The contribution of this paper can be concretely summarized into three folds as follows:

\begin{itemize}[leftmargin=4.5mm, itemsep=0.01em]
    \item We critically study the challenges associated with the lack of flexibility and omission of potentially impactful knowledge in the realm of prompt engineering for KG-Enhanced Large Language Models.

    \item We introduce KELP, an innovative approach aiming to capture potentially impactful knowledge and incorporate it into the prompts of LLMs via trained path-text encoding, with two coverage rules ensuring the flexibility of knowledge extraction. 

    \item Extensive experiments on Fact Verification and Question Answering (QA) datasets that encompass diverse graph reasoning patterns demonstrate the effectiveness of KELP.
\end{itemize}

\section{Problem Formulation}

In this section, we introduce the task of enhancing LLM performance with KGs. The KG is defined as $\mathcal{G}=(\mathcal{E}, \mathcal{R}, \mathcal{T})$, where $\mathcal{E}$ and $\mathcal{R}$ are sets of entities and relations, respectively, and $\mathcal{T}=\{(h,r,t)|h,t\in\mathcal{E},r\in\mathcal{R}\} $ represents the set of knowledge triplets, each contains a head entity $h$, a tail entity $t$, and a relation $r$. In addition, we have a pretrained large language model (LLM) denoted as $LM$. Given a question $q$ as the given task, if we denote all the entities contained in $q$ as a set $\mathcal{E}_q$, the goal is to utilize the background KG $\mathcal{G}$ as input prompts to support the generation of $LM$ based on question $q$.

\begin{figure*}
    \centering
\includegraphics[width=\textwidth]{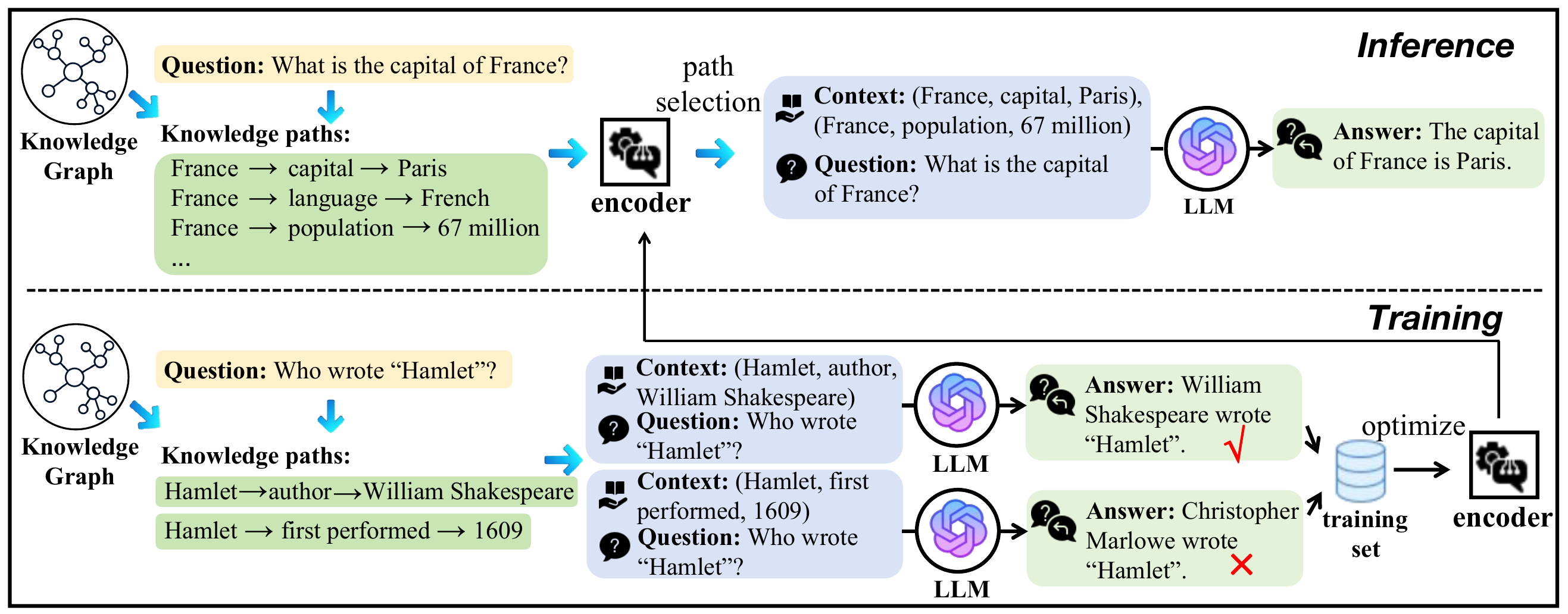}
    \caption{The overall pipeline of the proposed KELP.  During the inference phase, we identify knowledge paths from the knowledge graph that are associated with the entities present in the input question. An encoder is then trained to select valuable paths as knowledge contexts. Finally, the selected knowledge contexts, along with the input question, are input into the  LLM to generate the final answer. }
    \label{fig:pipeline}
\end{figure*}

\section{Methodology}

In this section, we introduce the details of our proposed framework KELP, which is presented in Figure~\ref{fig:pipeline}. KELP is structured into three phases: \textit{(i)} Knowledge path extraction, \textit{(ii)} Sample encoding, and \textit{(iii)} Fine-grained path selection. During knowledge path extraction, we extract a set of knowledge paths for each entity in the entity set $\mathcal{E}_q$ of $q$ from the background KG $\mathcal{G}$. For sample encoding, we employ a sentence encoder $M$ trained on the latent semantic space that can encode the input question $q$ and paths in the extracted knowledge path sets to obtain their distance (i.e., the possibility that the paths can influence the output of LLMs), thus ensuring capturing the potentially impactful knowledge in the paths. In the final fine-grained path selection phase, we propose two coverage rules to guarantee that the selection of knowledge paths is sufficiently flexible, thereby ensuring the acquisition of the most diverse and representative paths for the inference of LLMs regarding the input question $q$. Additionally, we also design an alternative strategy called \emph{Relation-Only Ranking} to generalize KELP to cases where the sizes of the knowledge path sets become substantially massive.


\subsection{Knowledge Path Extraction}
In this subsection, we introduce the knowledge path extraction in KELP. The objective is to identify valuable paths in the background Knowledge Graph $\mathcal{G}$, i.e., knowledge paths, that contain potentially impactful knowledge for a given input question $q$, which could be used as additional contexts in the prompt to improve the factual accuracy in the generation process of the LLM. To achieve this, we propose the following path extraction procedure: For each entity $e$ in the entity set $\mathcal{E}_q$, we first extract a knowledge path set denoted as follows:
\begin{equation}
\begin{aligned}
       & \mathcal{P}_e=\{(e\rightarrow r\rightarrow o)|o\in\mathcal{E}, r\in\mathcal{R}\}\cup\\ 
        &\{(e\rightarrow r_1\rightarrow o_1 \rightarrow r_2 \rightarrow o_2)|o_{1,2}\in\mathcal{E}, r_{1,2}\in\mathcal{R}\},
\end{aligned}
\end{equation}
which contains all 1-hop and 2-hop paths starting from the entity $e$. These selected paths serve as the candidates for the sample encoding phase. 

\subsection{Sample Encoding}
\label{sec:enc}
From the extracted path set $\mathcal{P}_e$, sample encoding aims to further refine the candidate knowledge paths that could help LLMs generate factually accurate answers for question $q$ via learned encoding. Specifically, we encode both the question $q$ and candidate knowledge paths in $\mathcal{P}_e$ via an encoder $M$ fine-tuned on latent semantic space. The fine-tuning steps of $M$ are introduced in Sections~\ref{opti1} and~\ref{opti2}. In this manner, we could quantify the usefulness of each path based on the learned representations obtained by $M$. 

To utilize the pretrained knowledge of the encoder $M$, we construct a path sentence for each knowledge path before encoding. The conversion depends on the number of triplets in the knowledge path: For a path containing only one triplet $(h,r,t)$, we formulate the path sentence $p'$ as $p' = ``h\ r\ t.$'' 
For a path consisting of two triplets $(h_1,r_1,t_1)$ and $(h_2,r_2,t_2)$, we construct the path sentence $p$ as follows: $p' = ``h_1\ r_1\ t_1,\ h_2\ r_2\ t_2.$'' The embeddings $\bh_q$ and $\bh_p$ for the question $q$ and the knowledge path $p$ is acquired by encoding $q$ and $p'$ using the encoder $M$ as follows: 
\begin{equation}
\bh_q = M(q), \ \ \ \ 
\bh_p = M(p').
\end{equation}
With the encoded representations $\bh_q$, $\bh_p$, we are ready to learn the beneficial paths that contain potentially impactful knowledge, based on the learned latent semantic similarity between $\bh_q$ and $\bh_p$. 

\subsection{Fine-Grained Path Selection}\label{path selection}
In this phase, we aim to select the most suitable paths as in-context facts for the input question $q$ based on the cosine similarity between their representations as scores with our proposed coverage rules. In this manner, we can address the challenge of rigid path selection by flexibly adjusting the hyperparameters within the coverage rules controlling the diversity and amount of the selected paths. Specifically, we aggregate the path sets of all entities in $\mathcal{E}_q$ as $\mathcal{P}_{q}=\bigcup_{e\in\mathcal{E}_q} \mathcal{P}_e \label{eq:all_path}.$ Notably, the paths in $\mathcal{P}_q$ inevitably involve redundant triplets that are shared in a larger number of paths. Therefore, we need to remove paths with overlapping triplets for selection. We first divide the entire path set $\mathcal{P}_q$ into subsets, each of which contains paths that share a specific triplet. By denoting the set of paths sharing a specific triplet $(h,r,t)$ as $\mathcal{P}_q(h,r,t)=\{p|(h\rightarrow r\rightarrow  t)\subset p,p\in\mathcal{P}_{q}\}$, we select the $k_1$ paths with highest scores for each triplet $(h,r,t)$ as follows:
\begin{equation}
\label{eq:k1}
\begin{aligned}
&\mathcal{P}’_q(h,r,t)= \argmax_{ \mathcal{P}’_q(h,r,t)} \sum\limits_{p\in\mathcal{P}’_q(h,r,t)}\cos(\bh_p, \bh_q), \\&\st\ \ |\mathcal{P}’_q{(h,r,t)}|=k_1, \ \ \mathcal{P}’_q{(h,r,t)}\subset \mathcal{P}_q(h,r,t).
\end{aligned}
\end{equation}
$\mathcal{P}’_q(h,r,t)$ represents the subset of $\mathcal{P}_q(h,r,t)$ with $k_1$-top paths in scores. By restricting the size of $\mathcal{P}’_q(h,r,t)$ to $k_1$, we could prevent including multiple high-scoring 2-hop triplets that share the same 1-hop triplet, precluding overly long contexts in the prompt with redundant information. We then select paths based on the scores from these subsets, where another rule is introduced to further restrict the number of distinct sharing triplets. Particularly, we denote the set of distinct sharing triplet subsets $\mathcal{T}’$ obtained as follows:
\begin{equation}
\label{eq:k2}
\begin{aligned}
\mathcal{T}’&=\argmax_{\mathcal{T}’}\sum\limits_{(h,r,t)\in\mathcal{T}’}\max_{p\in\mathcal{P}’_q(h,r,t)}\ \cos(\bh_p,\bh_q), \\
    &\st |\mathcal{T}’|\leq k_2.
    \end{aligned}
\end{equation}
Here, we introduce another parameter $k_2$ to control the size of $\mathcal{T}’$, which consists of the distinct sharing triplets that can constitute the aggregated path set:
\begin{equation}
        \mathcal{P}’_r=\bigcup_{(h,r,t)\in \mathcal{T}’} \mathcal{P}’_q(h,r,t), \\
\end{equation}
where $\mathcal{P}’_r$ is the aggregated path set from triplets in $\mathcal{T}’$.
By restricting the size of $\mathcal{T}'$, we can avoid the inclusion of excessive, irrelevant information in the context. Nonetheless, there still exist specific paths with low similarity to the input question $q$. Thus, we additionally consider a threshold to reduce the impact of such low-similarity paths. Particularly, we set the threshold as the lowest similarity score among the highest similarity scores among all selected $\mathcal{P}’_q(h,r,t)$, which is formally described as:
\begin{equation}
    \gamma=\min_{(h,r,t)\in\mathcal{T}’}\ \  \max_{p\in\mathcal{P}’_q(h,r,t)} \cos(\bh_p, \bh_q),
\end{equation}
where $(h,r,t)\in\mathcal{T}’$. In this manner, we could filter out the low-similarity paths in $\mathcal{P}’_r$ to obtain a high-similarity path set. The final selected reference path set is denoted as follows:
\begin{equation}
    \mathcal{P}_r=\{p|p\in\mathcal{P}’_r, \cos(\bh_p, \bh_q)\geq \gamma\}.
\end{equation}
As all paths in $\mathcal{P}_r$ are highly close to $q$, they will be selected as a context of the prompt fed to $LM$. 

By adjusting the value of $k_1$ and $k_2$ in Eqs. (\ref{eq:k1}) and (\ref{eq:k2}), we can flexibly control the path selection process in KELP and the amount of new knowledge introduced as context in the prompt, which allows for a more dynamic and tailored selection process, and ensures that the selected knowledge paths are optimally aligned with the knowledge required by $q$ to generate the desired outputs.


\subsection{Training-Set Establishment}\label{opti1}
To facilitate the training of encoder $M$ to match the candidate knowledge paths that can potentially improve the factual accuracy of the output of LLM, we construct a specialized dataset for encoder training, which contains both positive and negative instances that can or can not influence generation of the LLM. This dataset is constructed based on an original dataset comprising input questions and their corresponding ground-truth responses. Specifically, for a given input question $q$ that the Language Model $LM$ fails to generate the correct answer, we select the knowledge path set $\mathcal{P}_q$ as described in Section~\ref{path selection}. We select the knowledge paths in $\mathcal{P}_q$ and individually use each of them as the context for the LLM input. If the inclusion of a specific path results in the LLM generating the correct answer (i.e., the answer is consistent with ground truth), we consider this knowledge path as a positive sample of the training set. Similarly, we recognize it as a negative sample of the training set if it still leads to an incorrect answer.  

This training set is constructed using samples in the background KG and therefore encompasses a wide spectrum of reference paths that are both semantically related and semantically non-related to the input question $q$. This design ensures that our encoder $M$ is capable of learning the latent semantics that match both directly semantic-connected knowledge paths to the input question $q$, but also the potentially impactful knowledge paths that may not be (directly) semantically-related to the input question $q$, which substantially improves the generalization ability of the learned encoder in KELP. 


\subsection{Pairwise Optimization}\label{opti2}
With the positive and negative samples selected to establish the dataset, we proceed to train the sentence encoder $M$. During the training of the encoder $M$, we encode the input question $q$ and the corresponding path sentences converted from positive and negative samples (see subsection \ref{sec:enc}), denoted as $p_q^{+}$ and $p_q^{-}$. The representations of $q$, $p_q^{+}$ and $p_q^{-}$ are $\bh_q$, $\bh^{+}_q$ and $\bh^{-}_q$ respectively.

To train the encoder $M$, we design a pairwise loss with both the positive and the negative knowledge path samples for a given input question $q$. The loss function $\mathcal{L}$ is defined as follows:
\begin{equation}
\label{eq:loss}
\mathcal{L} = \sum_{q} \max(\cos(\bh_q,\bh^{-}_q) - \cos(\bh_q,\bh^{+}_q)+\eta, 0).
\end{equation}
Here, $\cos(\cdot,\cdot)$ represents the cosine similarity function, and $\eta$ is a threshold to prevent the model from excessively focusing on positive or negative samples. The loss function defined in Eq. (\ref{eq:loss}) encourages the embeddings of positive samples (where $q$ and the path $p$ are related) to be closer in similarity compared to the embeddings of negative samples (where they are not related), where the latent semantic learned by the encoder can be well aligned with the matchfulness between a potentially impactful knowledge path and an input question $q$. 
Through this optimization process, the model acquires the capability to capture useful knowledge that can enhance the output of LLMs, encompassing even potentially impactful knowledge that may not be immediately apparent or directly related.

\subsection{Relation-Only Ranking}
\label{sec:rel-rank}

The above training strategy can be well applied to the number of candidate path of normal KGs. However, in situations where the number of paths becomes substantially massive, we introduce an alternative path selection strategy called \emph{Relation-Only Ranking} to efficiently select important paths from the KG. This approach is particularly useful for large knowledge graphs where the $k$-hop $(k=1,2)$ neighboring subgraph of entities mentioned in the question tends to be excessively dense for path selection. In such cases, encoding every path we extract can be time-consuming. Recognizing that it is primarily the relations within the paths that provide the most valuable enhancements, we pivot to the \emph{Relation-Only Ranking} strategy.

In this strategy, when dealing with a specific extracted path $p$, we first construct the path sentence $p'$ exclusively from the relations present in $p$. This means that for a path comprising only one triplet $(h,r,t)$, we formulate a path sentence $p'$ as follows: $p' = ``r.$'' For a path containing two triplets $(h_1,r_1,t_1)$ and $(h_2,r_2,t_2)$, we construct a path sentence $p$ as follows: $p' = ``r_1, r_2.$'' This path sentence serves as the input to another encoder specifically designed for path sentences with only relations, denoted as $M_r$.

The encoder $M_r$ is trained in a similar manner as $M$, but it utilizes the new path sentences constructed solely from relations within the knowledge paths. Subsequently, we employ $M_r$ to rank the cosine similarity between the representations of input question $q$ and path sentence $p'$ containing relations, similar to how we generally rank knowledge paths. From the selected relations with high scores, we then use the original encoder $M$ to rank all the knowledge paths associated with these selected relations. The paths with higher scores are chosen as our contexts. With the introduced Relation-Only Ranking approach, we significantly reduce the number of candidate path sentences that require encoding in case of huge KG, resulting in a more efficient matching process that consumes less time while still being able to select valuable contexts based on the Relation-Only information.

\section{Experiments}
In this section, we introduce the extensive experiments conducted on two different tasks to demonstrate the effectiveness of the proposed method KELP. Our experimental setup closely follows the experimental settings outlined in KG-GPT~\cite{kggpt}, ensuring consistency and comparability with existing research in the field. 

\subsection{Datasets}
In this paper, we focus on two important tasks respectively on two different types of datasets for KG-Enhanced LLM: \textit{(i)} \textbf{Strongly Semantic Knowledge}, where the majority of questions have directly relevant semantic knowledge available in the KG, and \textit{(ii)} \textbf{Weakly Semantic Knowledge}, where only a minority of questions have directly relevant semantic knowledge accessible in the KG. For the Strongly Semantic Knowledge task, we utilize MetaQA~\cite{metaqa}, i.e., a crucial benchmark dataset containing subsets of questions with 1-hop/2-hop reasoning steps respectively, and featuring a wide variety of questions. Each question in MetaQA comes with a set of supporting facts and a corresponding query over a knowledge graph, challenging models to perform intricate multi-hop inference to derive accurate answers. With its rich contextual information, MetaQA presents significant challenges for models to effectively reason over interconnected entities and relations within the knowledge graph. 
For the Weakly Semantic Knowledge task, we utilize the FACTKG dataset~\cite{factkg}. The FACTKG dataset comprises 108,000 claims categorized as either \emph{True} or \emph{False}, where claims are subject to validation with DBpedia, i.e., a knowledge graph developed by ~\cite{dbpedia} which is not directly connected to most questions. In our experiments, we employ a subset of DBpedia provided by~\cite{factkg} in FACTKG.


\subsection{Baselines}
We include the same baselines as previous studies~\cite{kggpt}. We conduct all tasks with the large language model ``gpt-3.5-turbo-0613''~\cite{gpt}.

For all datasets, we conduct a question-only setting without evidence (i.e., context) in all few-shot learning scenarios. Moreover, for LLM-based evidence setting, where LLMs are utilized to capture the useful knowledge in KGs as prompts, we employ KG-GPT~\cite{kggpt} in the same settings to assess its performance across varying levels of evidence support from the KG. 
These experiments are designed to assess the model's performance across a spectrum of scenarios employing 4-shot, 8-shot, and 12-shot configurations. 

Besides few-shot learning settings, we also compare these performances with fully supervised models: For Strongly Semantic Knowledge setting, we implement five widely recognized baselines for Knowledge Graph Question Answering (KGQA), i.e., KV-Mem~\cite{kvmem}, GraftNet~\cite{graftnet}, EmbedKGQA~\cite{embedkgqa}, NSM~\cite{nsm}, and UniKGQA~\cite{unikgqa}. For Weakly Semantic Knowledge setting, we also compare these performances of few-shot learning settings with two encoder-only transformer-based text classifiers, BERT~\cite{bert} and  BlueBERT~\cite{bluebert}, and an evidence retrieval approach GEAR~\cite{gear}, which comprises an evidence graph retriever and a claim verification model. 


\begin{table}[!t]\centering
\caption{\label{resultmain}
Comparison between the accuracy(\%) and standard deviations over Few-shot settings of KELP and baselines on both tasks. Here LLME represents LLM-based evidence. $\Delta_{PE}$ is the improved value of KELP compared to LLME. \emph{Stongly} and \emph{Weakly} respectively represent Strong Semantic Knowledge and Weakly Semantic Knowledge. The best results in each learning strategy are shown in \textbf{bold}, respectively. 
}
\renewcommand{\arraystretch}{1.5}
\setlength\tabcolsep{5.0pt}
\small
\begin{tabular}{c|c|ccc|ccc|ccc}
\hline

\multicolumn{1}{c|}{\multirow{2}{*}{Task}} &
\multicolumn{1}{c|}{\multirow{2}{*}{Method}} & \multicolumn{3}{c|}{Accuracy (\%)} & \multicolumn{1}{c}{\multirow{2}{*}{$\sigma$}} \\ \cline{3-5}

& & \multicolumn{1}{c}{4-shot} & \multicolumn{1}{c}{8-shot} & 12-shot & \\ \hline

 & GPT& \multicolumn{1}{c}{54.4} & \multicolumn{1}{c}{61.5} & 63.0 & 4.59\\ 

Strongly&LLME & \multicolumn{1}{c}{94.7} & \multicolumn{1}{c}{95.8} & 96.3 & 0.82\\ 

1-hop&\cellcolor{gray!21}KELP & \cellcolor{gray!21}{\textbf{97.5}} & \cellcolor{gray!21}{\textbf{97.0}} & \cellcolor{gray!21}\textbf{97.1} & \cellcolor{gray!21}0.26 \\

&$\Delta_{PE}$ & \multicolumn{1}{c}{+2.8} & \multicolumn{1}{c}{+1.2} & +0.8 & -0.56\\ \hline

 & GPT& \multicolumn{1}{c}{22.6} & \multicolumn{1}{c}{22.8} & 28.3 & 3.23\\ 

Strongly&LLME  & \multicolumn{1}{c}{92.8} & \multicolumn{1}{c}{93.8} & \textbf{94.4} &  0.81\\ 

2-hop&\cellcolor{gray!21}KELP & \cellcolor{gray!21}{\textbf{93.7}} & \cellcolor{gray!21}{\textbf{94.3}} & \cellcolor{gray!21}93.8 & \cellcolor{gray!21}0.32 \\

&$\Delta_{PE}$ & \multicolumn{1}{c}{+0.9} & \multicolumn{1}{c}{+0.5} & -0.6 & -0.49\\ \hline

\multirow{4}{*}{Weakly} & GPT& \multicolumn{1}{c}{54.6} & \multicolumn{1}{c}{55.2} & 64.0 & 5.26\\ 

&LLME & \multicolumn{1}{c}{59.5} & \multicolumn{1}{c}{67.7} & \textbf{72.7} & 6.66\\ 

&\cellcolor{gray!21}KELP & \cellcolor{gray!21}{\textbf{68.5}} & \cellcolor{gray!21}{\textbf{68.6}} & \cellcolor{gray!21}69.2 & \cellcolor{gray!21}0.38 \\

&$\Delta_{PE}$ & \multicolumn{1}{c}{+9.0} & \multicolumn{1}{c}{+0.9} & -3.5 & -6.28\\ \hline
\end{tabular}
\end{table}

\subsection{Implementation Details}

In this subsection, we provide the detailed settings for the implementation of our framework. The baseline LLM used in our experiments is ``gpt-3.5-turbo-0613''~\cite{gpt}, which is an enhanced version within the GPT-3 series. During the establishment of the training set, we use 20\% of the samples from the original training sets to identify certain positive and negative triplets. Then, a pretrained DistilBert Model with 66 million parameters is introduced as the encoder $M$ to judge whether a triplet can potentially contain the important knowledge to correct/improve the generation of the baseline LLM. For optimization, we use AdamW~\cite{loshchilov2018decoupled} as the optimizer, with the learning rate set as $2\times 10^{-6}$. We set $k_1=k_2=4$ in the coverage rules based on their performance.

Here, for the FACTKG dataset, due to the large size of the neighboring subgraph associated with the entities, we employ the \emph{Relation-Only Ranking} strategy introduced in Section~\ref{sec:rel-rank} to select diverse and concise triplet paths from the KG. Furthermore, our preliminary analysis of the FACTKG dataset reveals notable accuracy in scenarios where a claim is determined to be \emph{True}. Conversely, in instances from the FACTKG dataset where a claim is predicted as \emph{False}, there exists a possibility that it could actually be a \emph{True} example, but with context information that has been inadequately captured. To mitigate this issue, for claims predicted to be \emph{False} within this dataset, we employ the LLM for a secondary verification process devoid of any contextual information.

\begin{figure*}
    \centering
    \includegraphics[width=1.0\linewidth]{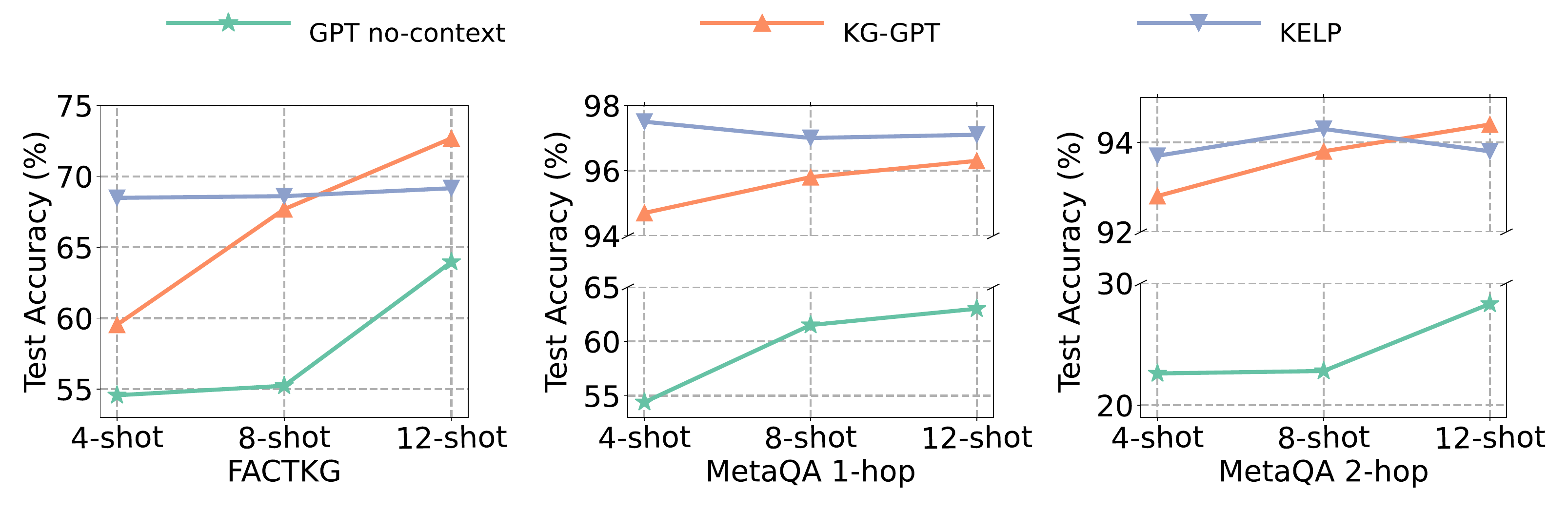}
    \caption{Comparison between the baseline GPT no-context, KG-GPT (LLM-based evidence), and our proposed method KELP on the FACTKG dataset and MetaQA dataset w.r.t. different shots in the learning setting.}
  \label{fig:shots}
\end{figure*}





\subsection{Results and Analysis}
In this subsection, we compare and analyze the performance of KELP and various baselines on the Strongly Semantic Knowledge and Weakly Semantic Knowledge tasks. The results of few-shot learning settings are summarized in Tables~\ref{resultmain}. From the table, we can find that adding LLM-identified useful knowledge in the prompt, i.e.,  LLM-based evidence, demonstrates significant performance improvement over the baseline GPT. This indicates that a lack of certain knowledge can indeed result in serious performance degradation in the LLM generation due to factual inaccuracy. However, since LLMs mainly encode direct semantic information, potentially useful knowledge with indirect semantic similarity with the input texts can be overlooked by LLM-based evidence. By finetuning pretrained encoder to capture the latent similarity between the collected sample pairs of impactful knowledge paths and input texts, in the experiments conducted within both 4-shot and 8-shot frameworks, our methodology obtains superior outcomes compared to those achieved through LLM-based evidence. Notably, in the 12-shot scenario, our approach's performance in 1-hop Strong Semantic Knowledge task with a retrieval surpassed that of LLM-based evidence. Furthermore, in other experimental settings within the 12-shot scenario, our method's results approached the efficacy levels of LLM-based evidence, demonstrating the potential of our approach to closely match or exceed LLM capabilities under varying conditions of informational support, especially in scenarios with a lower number of shots. The analysis of the relationship between performance and shots will be discussed in Section~\ref{ss}.

In table~\ref{resultfull} we provide the performances of fully supervised models.
Our research indicates that the method KELP we propose in few-shot learning settings surpasses the performance of some fully supervised models, achieving results that are close to the highest accuracy benchmarks among these models. This finding underscores the effectiveness of our approach in leveraging limited data to achieve high levels of accuracy.

\begin{table}[!t]\centering
\caption{\label{resultfull}
Baselines of fully supervised models on both tasks. \emph{Stongly} and \emph{Weakly} respectively represent Strong Semantic Knowledge and Weakly Semantic Knowledge. 
}
\renewcommand{\arraystretch}{1.5}
\setlength\tabcolsep{5.5pt}
\small
\begin{tabular}{c|c|cc}
\hline
\multirow{2}{*}{Semantic Knowledge} & \multirow{2}{*}{Methods} & \multicolumn{2}{c}{Accuracy  (\%) } \\ \cline{3-4}
 &  & \multicolumn{1}{c|}{1-hop} & 2-hop \\ \hline
\multirow{5}{*}{Strongly} & KV-Mem & \multicolumn{1}{c|}{96.2} & 82.7 \\ 
 & GraftNet & \multicolumn{1}{c|}{97.0} & 94.8 \\ 
 & EmbedKGQA & \multicolumn{1}{c|}{97.5} & 98.8 \\  
 & NSM & \multicolumn{1}{c|}{97.1} & 99.9 \\  
 & UniKGQA & \multicolumn{1}{c|}{97.5} & 99.0 \\ \hline
  & BERT & \multicolumn{2}{c}{65.20} \\  
 & BlueBERT & \multicolumn{2}{c}{59.93} \\  
\multirow{-3}{*}{Weakly} & GEAR & \multicolumn{2}{c}{77.65} \\ 
\hline
\end{tabular}
\end{table}



\subsection{Sensitivity w.r.t. Shots}\label{ss}

Furthermore, we compare KELP with the method on LLM-based evidence, when different shots of knowledge are included in the prompt. As the results illustrated in Figure~\ref{fig:shots} and the standard deviations in Table~\ref{resultmain}, the number of shots does not significantly affect the performance of KELP. This phenomenon can be attributed to KELP's emphasis on capturing potentially impactful knowledge aiming at effectively refining the outputs of LLM. In contrast, in-context examples serve merely to enhance the LLMs on a semantic level. Given that these in-context examples are crafted manually, it becomes challenging to determine the exact influence of increasing their number. Furthermore, the addition of more in-context examples can potentially introduce learning noise, detracting from the model's performance. 
Consequently, KELP exhibits minimal fluctuation in its performance across various numbers of shots, particularly excelling in contexts where the shots are limited. This stability and superior performance, even with scant examples, render KELP especially relevant in practical scenarios where acquiring a large volume of examples is challenging. The consistency of KELP under these conditions not only demonstrates its robustness but also its practical applicability, offering a compelling solution in environments where data limitations are a significant constraint.

\subsection{Ablation Study}
In this subsection, we design different variants of KELP to demonstrate the effectiveness of various components in our framework. In particular, we consider the following variants: (1) KELP w/o Ru1, which removes the $k_1$ in coverage rules and directly selects paths with the highest scores in each set $\mathcal{P}_q(h,r,t)$. As a result, we did not incorporate measures for fault tolerance regarding the selection of useful knowledge within the set. 
(2) KELP w/o Ru2, which removes the $k_2$ in coverage rules and directly selects the top-1 set with the highest scores. As a result, the diversity of selected paths cannot be ensured.
(3) KELP w/o Ra, which removes the ranking performed by the encoder and randomly selects paths in the subgraphs. As a result, the selected paths contain less beneficial information. 
The results of the ablation study, presented in Table~\ref{ablation}, validate the effectiveness of the coverage rules and the ranking method. Removing each of them will lead to a decrease in the accuracy of prediction, which demonstrates the effectiveness of our design.

\begin{table}[!t]\centering
\caption{\label{ablation}
Experimental results of ablation studies. \emph{Stongly} and \emph{Weakly} respectively represent Strong Semantic Knowledge and Weakly Semantic Knowledge.  Strongly 1-hop is not applicable to KELP w/o Ru1. 
}
\renewcommand{\arraystretch}{1.5}
\setlength\tabcolsep{2.8pt}
\small
\begin{tabular}{c|c|ccc}
\hline
\multirow{2}{*}{Tasks}&
\multirow{2}{*}{Methods} & \multicolumn{3}{c}{Accuracy (\%)} \\ \cline{3-5} 
& & \multicolumn{1}{c}{4-shot} & \multicolumn{1}{c}{8-shot} & 12-shot \\ \hline

\multirow{4}{*}{Strongly 1-hop}&
\cellcolor{gray!21}KELP & \cellcolor{gray!21}{\textbf{97.5}} & \cellcolor{gray!21}{\textbf{97.0}} & \cellcolor{gray!21}\textbf{97.1} \\
&KELP w/o Ru1 & \multicolumn{1}{c}{-} & \multicolumn{1}{c}{-} & - \\
&KELP w/o Ru2 & \multicolumn{1}{c}{90.0} & \multicolumn{1}{c}{89.4} & 89.6 \\
&KELP w/o Ra & \multicolumn{1}{c}{82.9} & \multicolumn{1}{c}{82.2} &  82.3\\ \hline

\multirow{4}{*}{Strongly 2-hop}&
\cellcolor{gray!21}KELP & \cellcolor{gray!21}{\textbf{93.7}} & \cellcolor{gray!21}{\textbf{94.3}} & \cellcolor{gray!21}\textbf{93.8} \\
&KELP w/o Ru1 & \multicolumn{1}{c}{88.1} & \multicolumn{1}{c}{88.2} & 88.2 \\
&KELP w/o Ru2 & \multicolumn{1}{c}{81.0} & \multicolumn{1}{c}{80.9} & 81.2 \\
&KELP w/o Ra & \multicolumn{1}{c}{70.4} & \multicolumn{1}{c}{70.1} &  69.7\\ \hline

\multirow{4}{*}{Weakly}&
\cellcolor{gray!21}KELP & \cellcolor{gray!21}{\textbf{68.5}} & \cellcolor{gray!21}{\textbf{68.5}} & \cellcolor{gray!21}\textbf{69.2} \\
&KELP w/o Ru1 & \multicolumn{1}{c}{67.0 } & \multicolumn{1}{c}{67.4} & 68.3 \\
&KELP w/o Ru2 & \multicolumn{1}{c}{66.5} & \multicolumn{1}{c}{67.0} & 67.7 \\
&KELP w/o Ra & \multicolumn{1}{c}{66.1} & \multicolumn{1}{c}{64.6} &  66.3\\ \hline
\end{tabular}
\vspace{-.12in}
\end{table}

\section{Related Work}
\subsection{Large Language Model (LLM)}

Large language model (LLM), such as GPT~\cite{gpt}, BERT~\cite{bert}, and T5~\cite{t5}, represents a pivotal development in natural language processing (NLP). Based on transformers \cite{vaswani2017attention} trained extensively on diverse datasets that encompass a wide spectrum of textual sources (including books, articles, and websites), LLMs acquire a profound understanding of semantics and reasoning ability in multiple languages.  In the era of LLMs, \textbf{prompt engineering} is a specialized technique to efficiently adapt pretrained LLMs to downstream tasks, aiming to elicit desired responses from these models by careful design and optimization of the input text presented to the LLM known as prompts for different tasks. Prompt engineering leverages the extensive training of LLMs on diverse datasets to guide and instruct them to generate specific outputs or perform particular tasks without laborious fine-tuning for each downstream task.

\subsection{Knowledge Graph-Enhanced LLM}

Knowledge Graphs (KGs) \cite{chen2020review} are organized repositories for knowledge structured as a collection of triplets $KG = \{(h, r, t)  \subseteq \mathcal{E} \times \mathcal{R} \times \mathcal{E}\}$. $\mathcal{E}$ and $\mathcal{R}$ represent the set of entities and relations, respectively. Each triplet in the KG consists of a head entity $h$, a relation $r$, and a tail entity $t$. KG-Enhanced LLM aims to use KG as external knowledge to support LLM generations~\cite{wang2023knowledge}. Two primary strategies are prevalent for integrating KG knowledge into LLMs: \textit{training phase enhancement} and \textit{inference phase enhancement}. The former involves embedding KG knowledge into LLMs via novel training objectives or directly incorporating KG data into input sequences~\cite{wang2022recognizing}, exemplified by models like ERNIE~\cite{ernie} and K-BERT~\cite{kbert}. However, these methods require extensive computational resources and frequent updates. Alternatively, the inference phase enhancement incorporates new knowledge through graph models or innovative prompt engineering, as seen in QA-GNN~\cite{qagnn} and KG-GPT~\cite{kggpt}. It is noteworthy that these methods offer a computationally efficient and flexible method for KG integration, where LLMs' reasoning capabilities can be enhanced without constant retraining.

\section{Conclusion}

In this paper, we present KELP, an innovative method to enhance Large Language Models (LLMs) with Knowledge Graphs (KG), aiming to flexibly capture potentially impactful knowledge that may lack direct semantic relevancy to the input texts. Specifically, we establish a training dataset with real examples of path-text pairs that demonstrate the correction of LLM outputs by including external knowledge as contexts. Subsequently, we train a path-text encoder to measure whether an instance of knowledge (represented as a given path in KG) contains potentially impactful knowledge for a specified input text. Based on the similarity score, two coverage rules are introduced to further refine the selected knowledge paths with high flexibility. Through experimental validation on two datasets, KELP has proven its superiority over other state-of-the-art baselines on KG-Enhanced LLMs. 

\section{Limitaions}

Our work performs path selection via an encoder trained on the latent semantic space. As introduced in Section~\ref{opti1}, to train an encoder proficient in capturing valuable knowledge contexts encompassing both direct and indirect semantic relationships, it is essential to construct a training set that covers a diverse spectrum of data types. Nevertheless, manually testing the paths surrounding entities within a text input via Large Language Models to discern and select positive and negative samples would be an exceedingly time-consuming process.

\section{Ethics Statement}
In our work, 
the knowledge in the background knowledge graph (KG) and the pretrained large language model (LLM) may involve information from raw data in the real world with social bias. Nevertheless, our method only selects knowledge path samples from KGs based on their relations to the input texts. Thus, as long as the input texts and samples do not preserve harmful information, we believe that our method does not present any negative social impacts.

\section{Acknowledgement}
This work is supported in part by the National Science Foundation under grants (IIS-2006844, IIS-2144209, IIS-2223769, CNS2154962, and BCS-2228534), the Commonwealth Cyber Initiative Awards under grants (VV-1Q23-007, HV2Q23-003, and VV-1Q24-011), the JP Morgan Chase Faculty Research Award, and the Cisco Faculty Research Award.

\bibliography{anthology}


\end{document}